\documentclass[10pt,twocolumn,letterpaper]{article}

\usepackage[pagenumbers]{cvpr} 
\usepackage{booktabs} 
\usepackage{pifont}
\usepackage{float}
\usepackage{tikz}
\usepackage{lipsum} 
\usepackage[ruled]{algorithm2e} 
\usepackage{makecell}
\usepackage{bm}
\usepackage{dsfont}
\usepackage{graphicx}
\usepackage{ulem}
\usepackage{mathrsfs} 
\usepackage{multirow}

\usepackage{colortbl} 
\usepackage{soul} 
\usepackage[accsupp]{axessibility}

\newcommand{\supmatCOLOR}{black}

\newcommand{\supmat}{\textcolor{\supmatCOLOR}{{Sup.~Mat.}}\xspace}

\definecolor{first}{RGB}{255,178,178}
\definecolor{second}{RGB}{255,217,178}
\definecolor{third}{RGB}{255,255,178}
\definecolor{ours}{RGB}{220, 220, 220}
\providecommand{\best}{\cellcolor{red!20}}   
\providecommand{\second}{\cellcolor{orange!20}} 
\providecommand{\ours}{\cellcolor{gray!20}} 

\newcommand{\methodname}{VHOI}

\newcommand{\textprompt}{\tau}
\newcommand{\inputimg}{I}
\newcommand{\outputvideo}{V}
\newcommand{\height}{H}
\newcommand{\width}{W}
\newcommand{\heightlatent}{H'}
\newcommand{\widthlatent}{W'}
\newcommand{\numfeaturechannels}{C}

\newcommand{\numframes}{T}
\newcommand{\numframeslatent}{T'}

\newcommand{\diffusionindex}{i}

\newcommand{\traj}{\xi}
\newcommand{\numtrajectories}{K}
\newcommand{\inputobjectmask}{M_o} 
\newcommand{\inputhumanmask}{M_h} 

\newcommand{\humanmaskseq}{\mathbf{M}_h}        
\newcommand{\objectmaskseq}{\mathbf{M}_o}       
\newcommand{\inputhoimask}{\mathbf{M}_\text{hoi}^0} 
\newcommand{\hoimaskseq}{\mathbf{M}_\text{hoi}} 
\newcommand{\hoimaskseqpred}{\hat{\mathbf{M}}_\text{hoi}} 
\newcommand{\trajvideo}{M_\traj}
\newcommand{\vistraj}{\mathbf{o}}
\newcommand{\vismask}{M_v}
\newcommand{\trajvideoblur}{\tilde{M}_\traj}
\newcommand{\vismaskblur}{\tilde{M}_v}
\newcommand{\gaussiansmooth}{\sigma}
\newcommand{\kernelradius}{k}

\newcommand{\rgblatent}{l_\text{img}}
\newcommand{\videolatent}{l_{v}}

\newcommand{\trajfeature}{F_{a}}
\newcommand{\visibilityfeature}{F_{v}}
\newcommand{\hoifeature}{F_\text{hoi}}
\newcommand{\hoifeaturegated}{\tilde{F}_\text{hoi}}
\newcommand{\trajaugmentor}{\boldsymbol{\mathcal{A}}}
\newcommand{\densemodel}{\boldsymbol{\mathcal{D}}}
\newcommand{\basemodel}{\boldsymbol{\mathcal{B}}}

\newcommand{\trajextractor}{\mathcal{E}_t}

\newcommand{\patchemb}{\mathcal{E}_p}
\newcommand{\patchembaug}{\mathcal{E}_\text{pa}}
\newcommand{\visemb}{\mathcal{E}_v}
\newcommand{\hoiextractor}{\mathcal{E}_\text{hoi}}
\newcommand{\confidence}{\mathbf{s}}

\newcommand{\augmentorfuser}{\mathcal{E}_a}
\newcommand{\maskfuser}{\mathcal{E}_m}
\newcommand{\confidencehead}{\mathcal{E}_c}
\newcommand{\augscale}{\gamma_{\text{aug}}}
\newcommand{\augshift}{\beta_{\text{aug}}}
\newcommand{\augscalegated}{\tilde{\gamma}_{\text{aug}}}
\newcommand{\augshiftgated}{\tilde{\beta}_{\text{aug}}}
\newcommand{\gatevis}{G_v}
\newcommand{\dittoken}{F_\text{dit}}
\newcommand{\dittokenmodulated}{\tilde{F}_\text{dit}}
\newcommand{\diffloss}{L}

\definecolor{cvprblue}{rgb}{0.21,0.49,0.74}
\usepackage[pagebackref,breaklinks,colorlinks,allcolors=cvprblue]{hyperref}

\title{\methodname: Controllable Video Generation of Human-Object Interactions from Sparse Trajectories via Motion Densification}

\author{
Wanyue Zhang\textsuperscript{1}\quad
Lin Geng Foo\textsuperscript{1}\quad
Thabo Beeler\textsuperscript{3}~~~~~~~
Rishabh Dabral\textsuperscript{1,2}\quad
Christian Theobalt\textsuperscript{1,2}
\smallskip\\
\textsuperscript{1}MPI for Informatics, SIC\quad\quad
\textsuperscript{2}VIA Center\quad\quad
\textsuperscript{3}Google 
}

\begin{document}
\twocolumn[{
\renewcommand\twocolumn[1][]{#1}
\maketitle
\begin{center}
    \captionsetup{type=figure}
    \vspace{-5mm}
    \includegraphics[width=0.88\textwidth]{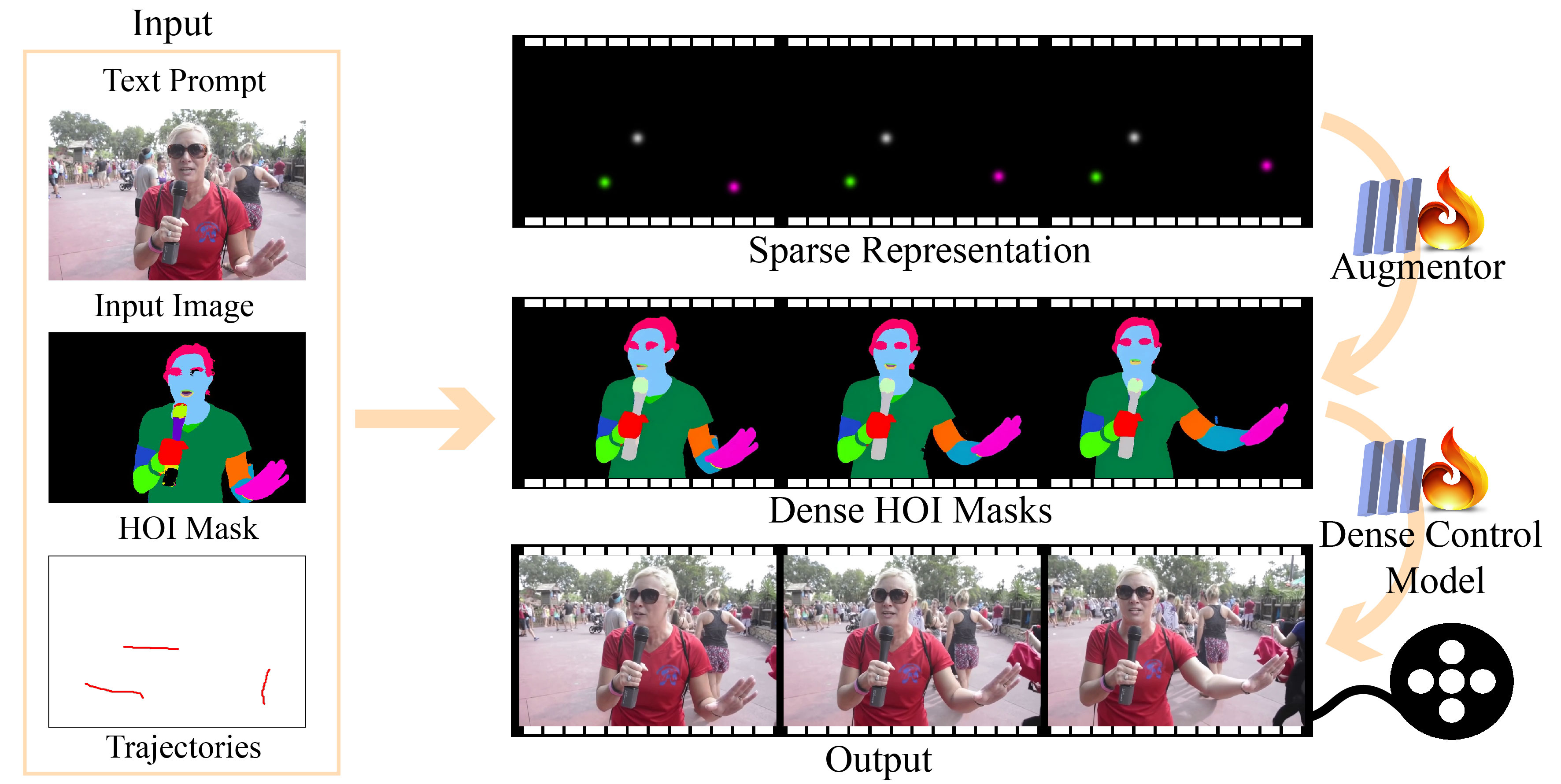}
	\caption
	{
    The input to \methodname{} consists of a text prompt, an input image, an HOI mask image, and trajectories. We first convert the trajectories into sparse representations (row 1) and learn an augmentor that generates dense HOI masks (row 2) from these sparse inputs. The resulting dense HOI masks are then used to control the video generation process. In our proposed sparse-to-dense representations, different colors denote different human parts and objects, providing fine-grained grounding of human-object interactions. This sparse-to-dense paradigm enables realistic HOI generation with easily specifiable, yet instance-aware controllability.
	}
    \label{fig:teaser}
\end{center}
}]

\begin{abstract}
Synthesizing realistic human-object interactions (HOI) in video is challenging due to the complex, instance-specific interaction dynamics of both humans and objects. Incorporating controllability in video generation further adds to the complexity. Existing controllable video generation approaches face a trade-off: sparse controls like keypoint trajectories are easy to specify but lack instance-awareness, while dense signals such as optical flow, depths or 3D meshes are informative but costly to obtain. We propose \methodname, a two-stage framework that first densifies sparse trajectories into HOI mask sequences, and then fine-tunes a video diffusion model conditioned on these dense masks. We introduce a novel HOI-aware motion representation that uses color encodings to distinguish not only human and object motion, but also body-part-specific dynamics. This design incorporates a human prior into the conditioning signal and strengthens the model’s ability to understand and generate realistic HOI dynamics. Experiments demonstrate state-of-the-art results in controllable HOI video generation. \methodname{} is not limited to interaction-only scenarios and can also generate full human navigation leading up to object interactions in an end-to-end manner.
Project page: \href{https://vcai.mpi-inf.mpg.de/projects/vhoi/}{https://vcai.mpi-inf.mpg.de/projects/vhoi/}
\end{abstract}
    
\vspace{-5mm}
\section{Introduction}
\label{sec:intro}
Human-object Interaction (HOI) video generation aims to synthesize realistic videos of people manipulating objects, for example, pulling a chair or opening a door~\cite{huang2025hunyuanvideo, zhang2025tora2, kim2025target, li2025genhsi, akkerman2025interdyn}. 
Understanding human-object interactions has been a key focus of the recent human-centric vision research.
While significant research efforts have been devoted towards HOI perception (reconstruction, detection, classification, etc.~\cite{aytekin2025follow, liu2022hoi4d, wang2025magichoi, fan2024hold, gkioxari2018detecting, narasimhaswamy2020detecting, xie2023category, lighten, betsubetsu}), only recently has 3D and 2D HOI synthesis garnered the attention of the community.
HOI scenes are particularly difficult because they require fine-grained understanding of hand articulation, contact, and object motion, which generic video models struggle to capture.
Controllable HOI video generation allows users to guide interactions using motion cues such as trajectories~\cite{zhang2025tora2} and masks~\cite{akkerman2025interdyn, li2025mask2iv, kim2025target}, with potential applications in streamlining animation workflows and generating synthetic training scenarios for robotics.
\par
However, controllability introduces several challenges depending on the control signal. 
\textit{Sparse} motion signals, such as 2D trajectories, are intuitive for the end-users but lack sufficient spatial and geometric structure, often leading to incoherent or physically implausible synthesis of interactions. 
For example, when a hand trajectory directs a grasping motion (as shown in~\cref{fig:teaser}), it is challenging for the state-of-the art models to move the arm, the body, and the object coherently. 
On the other hand, \textit{dense} signals, such as per-frame masks, depth maps, or 3D meshes, provide stronger motion cues, but are costly and impractical to obtain at inference.
In this work, we present a general approach to integrate the complementary benefits of sparse and dense control signals to improve HOI generations.
Given a text prompt, an input image, the initial HOI mask, and user-defined trajectories, our objective is to synthesize realistic human-object interactions.
To achieve it, we train an Augmentor network that converts the sparse trajectories into dense mask sequences for both the human and the object, providing rich motion cues to the video model. 
We then fine-tune a base video model on these densified/augmented control signals, enabling realistic and coherent HOI motion, as shown in~\cref{fig:teaser}.
\par
Naturally, the central question in our design is how to represent HOI motion signals and what intermediate form they should take. 
We propose to enrich motion signals with HOI semantics for both the sparse and dense training stages. 
For sparse trajectories, we color trajectory tracks using a fixed part-based palette derived from a human foundation model~\cite{khirodkar2024sapiens}, along with a distinct color for object motion. 
During dense conditioning, part-based human masks follow the same consistent color palette, ensuring that semantic meaning remains stable across frames, viewpoints and camera motion. 
This design injects a strong human prior into the conditioning signal and improves the model’s ability to understand and generate coordinated human-object motion.
Interestingly, we found that this approach of synthesizing videos from \textit{densified} sparse signals is versatile in that it also produces consistent improvements with other control signals like optical flow or instance masks.
In summary, our main contributions are threefold:
\begin{itemize}
\item We demonstrate the importance of bridging sparse motion inputs to denser, structured intermediate representations, and introduce a two-stage framework for controllable HOI video generation that achieves this through a learned motion augmentor.
\item We deviate from the conventional approaches for sparse motion control and propose novel HOI-aware motion representations which enhance instance-awareness and improves interaction fidelity.
\item We conduct extensive experiments across diverse HOI scenarios, including hand-object manipulation and full-body navigation-interaction sequences. Experimental results demonstrate that our method consistently improves controllability and realism over state-of-the-art baselines.
\end{itemize}
\section{Related Work}
\label{sec:related_work}
\subsection{Controllable Video Generation}
State-of-the-art video models~\cite{liu2024sora, yang2024cogvideox, hong2022cogvideo, kong2024hunyuanvideo, wan2025, chen2025goku} typically support conditioning on text prompts and input images; however, they lack finer-grained control.
Existing approaches to achieve controllability can be broadly categorized into three types:
Training-based methods, which retrain the entire base model with additional modalities~\cite{alhaija2025cosmos} or modified noise warping strategies, often at significant computational cost~\cite{burgert2025go};
(2) Fine-tuning approaches, leveraging lightweight modules such as ControlNet~\cite{li2024controllable, tu2025videoanydoor, geng2025motion}, LoRA~\cite{hu2022lora, zhang2026controllable}, ReferenceNet~\cite{hu2024animate}, or customized feature adapters~\cite{zhang2025tora, xiao2025trajectory, zhang2025tora2, mou2024t2i,peng2024controlnext} while keeping the base model frozen; and
3) Training-free approaches~\cite{qiu2024freetraj, xiao2024video}, which manipulate the noise injection or attention~\cite{ma2023trailblazer}, and may rely on test-time optimization~\cite{xiao2024video, foo2026physical}.
The control signals used in these methods can include motion trajectories, camera movement, scene layout, or event sequences~\cite{wang2023videocomposer, alhaija2025cosmos, hou2025training, huang2022layered, valevski2024diffusionmodelsrealtimegame, jin2025flovd, wu2025mind, wang2025cinemaster}. 
In the following section, we focus on motion-specific control signals.

\subsection{Motion-guided Video Generation}
Motion signals take various forms such as point-wise trajectories~\cite{zhang2025tora, wang2025ati,fu20243dtrajmaster, tanveer2025multicoin}, bounding boxes~\cite{ma2023trailblazer, wei2024dreamvideo, xing2025motioncanvas}, optical flow~\cite{shi2024motion, jin2025flovd}, masks~\cite{li2025mask2iv, akkerman2025interdyn, yariv2025throughthemaskmaskbasedmotiontrajectories}, edges~\cite{alhaija2025cosmos}, sketches~\cite{wang2023videocomposer}, human poses~\cite{hu2024animate, hu2025animateanyone2}, 3D human parameterizations~\cite{zhu2024champ, men2024mimo, li2025multimodal, shao2024isa4d, zhang2026controllable} or motion features from the reference videos~\cite{xiao2024video}.
Dense signals such as depth maps~\cite{alhaija2025cosmos} and 3D meshes~\cite{shao2024isa4d} offer more precise controllability and higher video quality, but they are difficult to obtain at inference time, which limits their practicality for user-specified content generation.
Several works have attempted to bridge the gap between sparse and dense signals by training separate networks to predict masks~\cite{yariv2025throughthemaskmaskbasedmotiontrajectories} or optical flows~\cite{shi2024motion, foo2026physical} from sparse trajectories.
Motion Prompting~\cite{geng2025motion} expands user-specified mouse clicks by leveraging a grid of motion tracks, and has showed applications in both object movement and camera control.
However, these methods are designed for generic video generation and do not capture HOI-specific semantics.
Our work investigates the key question of what constitutes a suitable intermediate motion representation for HOI, bridging the gap between sparse trajectories and fine-grained HOI masks with part-level semantics.
\subsection{Human Object Interaction Synthesis}
Human-object interaction (HOI) is a rapidly evolving field encompassing data collection~\cite{fan2023arctic, Jiang2022FullBodyHOI,liu2025core4d, liu2022hoi4d}, understanding~\cite{gkioxari2018detecting, ulutan2020vsgnet, ning2023hoiclip, liao2022gen, lighten}, reconstruction~\cite{xie2022chore, aytekin2025follow, Wang_2025_CVPR, huo2024monocular, wang2025magichoi, betsubetsu}, and synthesis in both 2D~\cite{Hu_2025_CVPR, ye2023affordance, hu2022hand} and 3D~\cite{ye2024g, Zhang_2025_CVPR,zhou2024gears, li2023object,li2024controllable,wu2024human,zhang2025bimart,yang2020roam,zhang2022couch,hassan2021stochastic,starke2019neural,liu2025core4d,he2024syncdiff,liu2024mimicking,xu2023interdiff,kulkarni2024nifty,yi2024generating,li2024zerohsi,jain2009interactive,he2025syncdiff, ghosh2022imos}.
There has been growing interest in HOI video synthesis~\cite{li2025genhsi, kim2025target, akkerman2025interdyn, huang2025hunyuanvideo}, driven by the rapid progress of high-quality large video generation models~\cite{kong2024hunyuanvideo, hong2022cogvideo, yang2024cogvideox,chen2025goku, wan2025, liu2024sora} and large-scale datasets~\cite{liu2025hoigen}.
Among HOI video synthesis methods, TAVID~\cite{kim2025target} introduces an attention-based loss to select target objects for interaction from input masks, but cannot generate navigation and interaction jointly without relying on an external interpolation model.
GenHSI~\cite{li2025genhsi} decomposes the task into script writing, 3D keyframe generation and frame interpolation, where keyframe generation requires costly inpainting and pose optimization.
InterDyn~\cite{akkerman2025interdyn} instead uses ground-truth mask sequences as control signals, which are difficult to obtain in practice.
In contrast, \methodname{} bridges the gap between sparse and dense control signals, achieving end-to-end navigation and interaction generation without external models.

\section{Method}
\label{sec:method}

Given an input image $\inputimg \in \mathbb{R}^{\height \times \width \times 3}$, a text prompt $\textprompt$, a set of $\numtrajectories$ user-specified point trajectories $\traj \in \mathbb{R}^{\numtrajectories \times \numframes \times 2}$ with optional frame-wise visibilities $\vistraj \in \mathbb{R}^{\numtrajectories \times \numframes} \in \{0, 1\}$, and the first-frame masks for the human $\inputhumanmask$ and the object $\inputobjectmask$, our goal is to synthesize a realistic HOI video $\outputvideo \in \mathbb{R}^{\numframes \times \height \times \width \times 3}$ conditioned on these inputs.
Here, $\numframes$ denotes the number of frames and $(\height, \width)$ the spatial resolution.
We assume that $\inputimg$ contains both the human (or hand) and the target object; thus, $\inputhumanmask$ and $\inputobjectmask$ can be obtained using off-the-shelf segmentation methods~\cite{ren2024grounded, khirodkar2024sapiens}.

Motivated by the importance of instance-aware motion cues for realistic HOI synthesis, \methodname{} consists of (1) a trajectory augmentor $\trajaugmentor$ that converts 
sparse trajectories $\traj$ into dense HOI mask sequences 
$\hoimaskseq \in \mathbb{R}^{\numframes \times \height \times \width \times 3}$ 
as an intermediate motion representation, and (2) a dense control model $\densemodel$ that generates the video conditioned on $\hoimaskseq$.
\cref{fig:method_sparse} and \cref{fig:method_dense} depicts an overview of the framework. 

In the following sections, we first review the preliminaries for video diffusion models in \cref{sec:prelim}, then introduce the proposed HOI-aware motion representations used for sparse and dense control signals in \cref{sec:motion_rep}. We next describe the trajectory augmentor in \cref{sec:traj_aug}, followed by the dense control model in \cref{sec:dense_model}.

\subsection{Video Diffusion Model Preliminaries}
\label{sec:prelim}
\begin{figure*}[!hbt]
 \centering
	\includegraphics[width=1\linewidth]{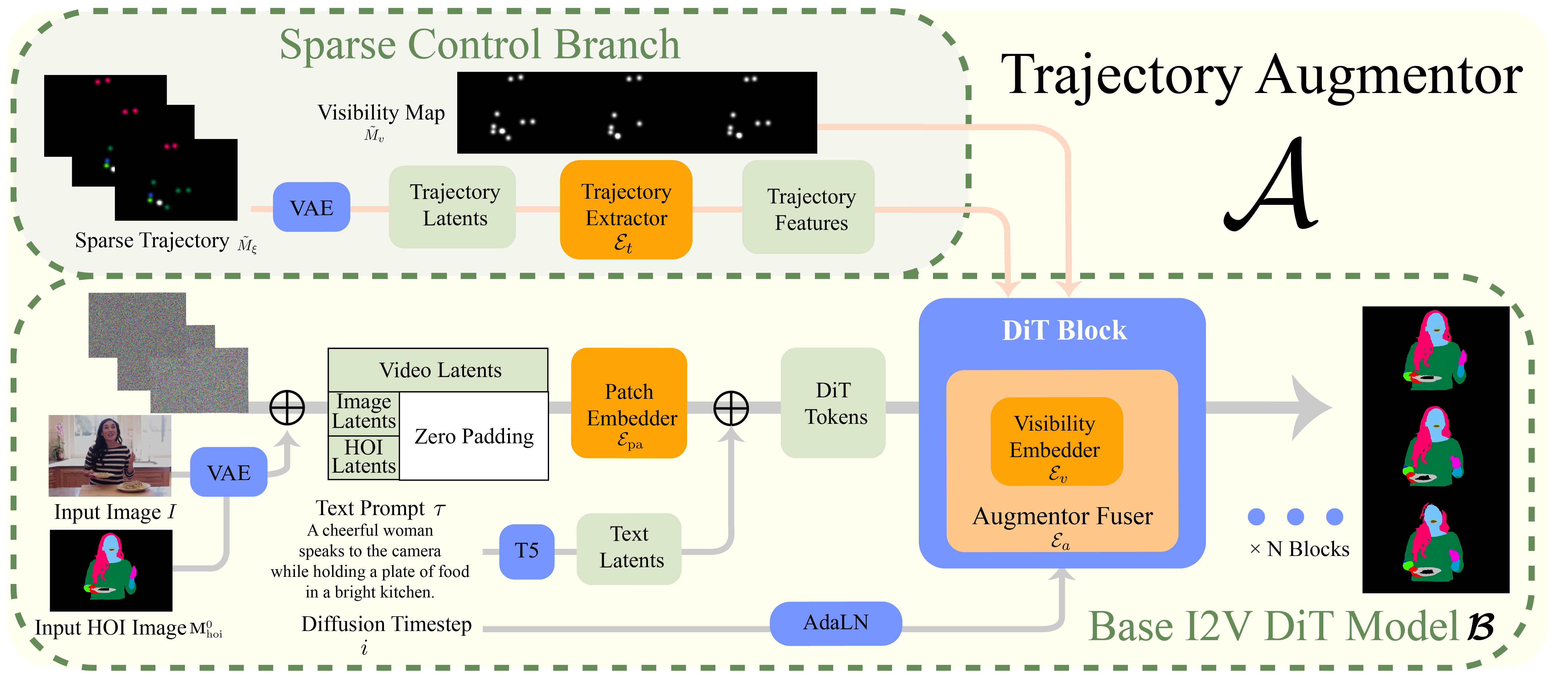}
        \caption
	{
The trajectory augmentor $\trajaugmentor$ receives sparse trajectories and the corresponding visibility maps (optional) as inputs. 
The trajectories are processed by a trajectory extractor and fused with transformer latents and visibility cues in the augmentor fuser, producing a sequence of HOI masks that densifies the sparse control signals, used in the dense control model $\densemodel$ as shown in \cref{fig:method_dense}.
Orange modules denote learnable components; blue modules are frozen.
} 
	\label{fig:method_sparse}
\end{figure*}

Our controllable video generation method builds upon the DiT-based~\cite{peebles2023scalable} image-to-video latent diffusion model from CogVideoX~\cite{hong2022cogvideo}.
A 3D VAE encodes the conditioning image $\inputimg$ into image latents
$\rgblatent \in \mathbb{R}^{1 \times \heightlatent \times \widthlatent \times \numfeaturechannels}$ and the target video $\outputvideo$ into video latents
$\videolatent \in \mathbb{R}^{\numframeslatent \times \heightlatent \times
\widthlatent \times \numfeaturechannels}$. 
This ensures that both modalities share the same spatial resolution and channel dimensionality
in the latent space.
The image latent, $\rgblatent$, is zero-padded along the temporal dimension, and concatenated
with the noisy video latents $\videolatent$ along the channel axis, and
fed into a patch embedder $\patchemb$ to produce a sequence of video
tokens. 
These tokens are then prepended with text tokens to form the combined token sequence $\dittoken$.
Hybrid attention is performed over $\dittoken$, and the DiT blocks rely on AdaLN~\cite{peebles2023scalable} to inject the diffusion timestep $\diffusionindex$.
Please see \cref{fig:method_sparse} and \cref{fig:method_dense} for the detailed architecture.

Training follows the velocity-prediction objective:
\begin{equation} \label{eq:diffusion_loss} \diffloss = \mathbb{E}_{\videolatent, \diffusionindex, \boldsymbol{\epsilon}}\Big[\big\| \basemodel(\dittoken^{\diffusionindex}, \mathbf{c}) - \mathbf{v}(\videolatent, \boldsymbol{\epsilon}, \diffusionindex) \big\|_2^2\Big], \end{equation} 
where $\boldsymbol{\epsilon} \sim \mathcal{N}(0, \mathbf{I})$, $\mathbf{c} = \{\textprompt, \traj\}$, $\basemodel$ is the DiT backbone, and $\mathbf{v}(\videolatent, \boldsymbol{\epsilon}, \diffusionindex)$ is the target velocity.
In \methodname{}, we freeze the base I2V DiT model $\basemodel$, introduce the HOI-aware motion features for controllability (\cref{sec:motion_rep}), and then fine-tune the additional feature extractors and fusers (\cref{sec:traj_aug} and \cref{sec:dense_model}).
\subsection{Sparse to Dense Motion Representations}
\label{sec:motion_rep}
Prior work has explored learning dense optical flow from sparse trajectories~\cite{shi2024motion}, but such flow lacks HOI semantics and can be distorted by camera motion rather than foreground movement. 
Using distinct colors per instance mask is another option~\cite{yariv2025throughthemaskmaskbasedmotiontrajectories}, yet it does not provide fine-grained control over human-object interactions.

\begin{figure}[t]
	\includegraphics[width=\linewidth]{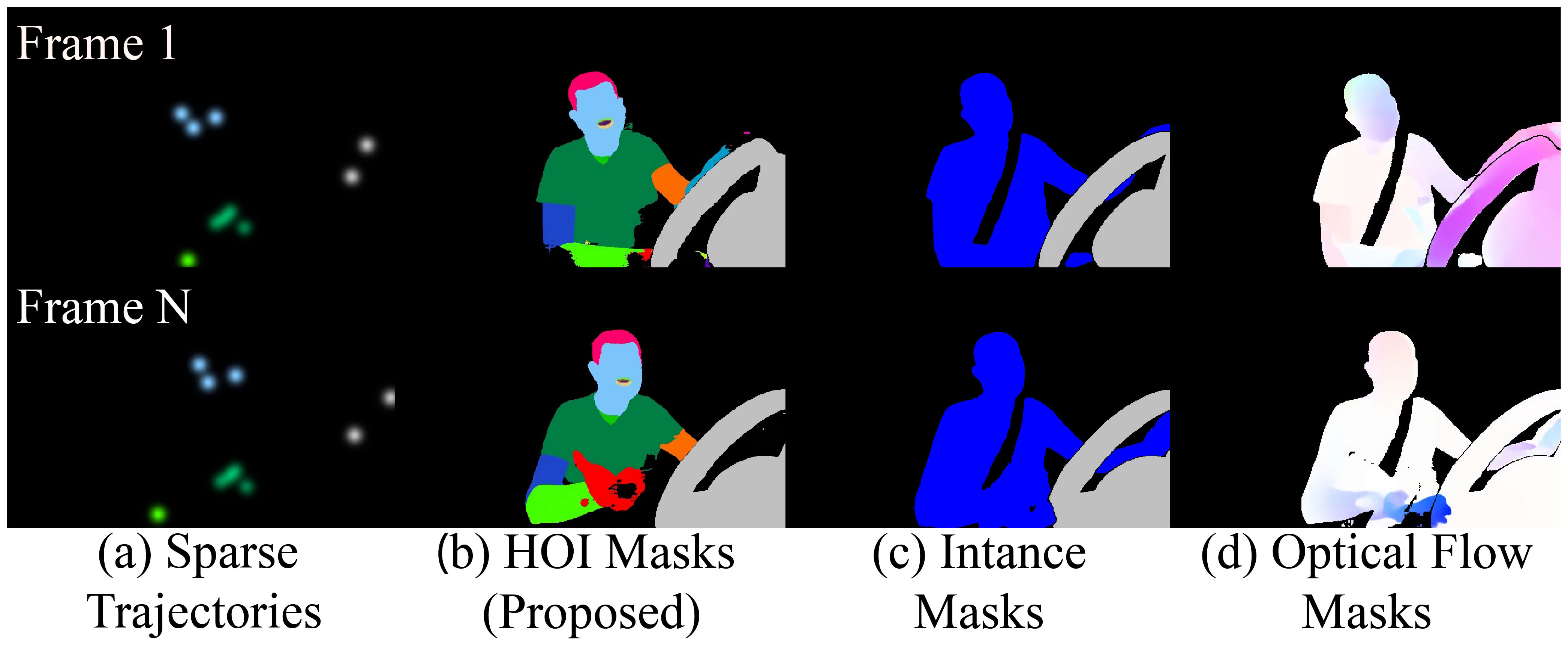}
	\caption{
        \textbf{Motion Representation}. We visualize two frames of colored sparse trajectories alongside the three intermediate motion representations studied in this work.
(a) The sparse trajectory representation, where different colors denote different human parts or objects.
(b) HOI masks (ours): constructed by combining object masks~\cite{ren2024grounded} with part-level human segmentation~\cite{khirodkar2024sapiens}, each assigned a consistent color to encode fine-grained HOI semantics.
(c) Instance masks: a coarser alternative that distinguishes only human and object regions, lacking part-level detail and interaction awareness.
(d) Foreground optical flow: computed via RAFT~\cite{teed2020raft} and masked to foreground regions; while the color encoding reflects motion magnitude and direction, it does not convey part-level or HOI-specific semantics.
        }
	\label{fig:motion_rep}
\end{figure}

\methodname{} relies on two levels of motion representations:  
(1) a sparse representation that encodes the trajectories $\traj$; and
(2) a learnable intermediate HOI mask sequence $\hoimaskseq$ that densifies $\traj$ and lowers the generation complexity, as shown in~\cref{fig:motion_rep}. 
We inject HOI semantics into both representations to enable instance-aware interaction reasoning. 

\textbf{Sparse representation.}
Given a user-specified sparse trajectory $\traj \in \mathbb{R}^{\numframes \times 2}$, the conventional approach draws the 2D path on a canvas~\cite{wang2025ati}, producing a video tensor $\trajvideo \in \mathbb{R}^{\numframes \times \height \times \width \times 3}$ that is then encoded by a VAE into the latent space of the video model.
Prior works~\cite{zhang2025tora, zhang2025tora2} also employ optical-flow color encoding~\cite{baker2011database} to indicate motion direction, but such representations do not encode part-aware HOI semantics.
Instead, we color-code the trajectory tracks using a fixed part-based palette, where each human body part and the target object receives a consistent color across scenes and motions (see~\cref{fig:motion_rep}(b)). 
This representation is more fine-grained than human–object coloring and provides explicit spatiotemporal cues linking body parts to objects.
If available, our model also allows incorporating the track visibilities in the form of a binary visibility map $\vismask \in \{0,1\}^{\numframes \times \height \times \width}$.
This optional signal can further help the model to modulate trust in the sparse control signal.
To ensure the trajectory and visibility signals remain visible after VAE encoding, we apply Gaussian blur to both $\trajvideo$ and $\vismask$ following~\cite{zhang2025tora}.
The blurred trajectory $\trajvideoblur$ and visibility map $\vismaskblur$ serve as inputs to a trajectory augmentor $\trajaugmentor$, which predicts the dense HOI mask sequence $\hoimaskseq$.
More information about the color palette and tracking visibility is included in \supmat.
\textbf{Dense Representation.}
Sparse trajectories convey high-level motion intent and are intuitive for users to specify, but they do not encode how humans and objects deform or interact over time.
To provide structured guidance signals, we introduce a dense motion representation built from ground-truth human and object mask sequences.
Specifically, we combine the human and the object mask sequences
$\humanmaskseq, \objectmaskseq \in \mathbb{R}^{\numframes \times \height \times \width \times 3}$
into a sequence of HOI masks $\hoimaskseq \in \mathbb{R}^{\numframes \times \height \times \width \times 3}$ via an element-wise maximum operation:
\begin{equation}
\begin{aligned}
\hoimaskseq^{(t)}(u,v) &=
\max\big( \humanmaskseq^{(t)}(u,v), \objectmaskseq^{(t)}(u,v) \big), \\
 t&=1,\dots,\numframes.
 \end{aligned}
\label{eq:mask_overlay}
\end{equation}
\cref{fig:motion_rep}(a) illustrates the output of this operation.
The color encoding of $\hoimaskseq$ is consistent with the sparse trajectory rendering $\trajvideo$, thereby preserving fine-grained HOI semantics across representations.
These dense masks explicitly model how body parts and the object deform and interact over time, serving as a rich intermediate representation that encodes strong instance-aware motion cues.

Once the augmentor $\trajaugmentor$ predicts $\hoimaskseq$, we feed it to the dense control model $\densemodel$ to synthesize $\outputvideo$.
\cref{sec:traj_aug} and \cref{sec:dense_model} detail $\trajaugmentor$ and $\densemodel$ respectively.
%

\subsection{Trajectory Augmentor}
\label{sec:traj_aug}
Given sparse motion cues $\trajvideoblur$ and $\vismask$ derived from pointwise trajectories $\traj$, 
a straightforward solution would be to directly condition an I2V diffusion model $\basemodel$ by finetuning a control branch. 
However, such sparse signals do not encode instance boundaries or the spatial motion extent. 
Consequently, the model struggles to produce coherent HOI motion.
Our key idea is to train an augmentor $\trajaugmentor$ that transforms sparse cues $(\trajvideoblur, \vismask)$ into structured HOI masks $\hoimaskseq$, providing richer instance-level motion cues for interaction synthesis.
$\trajaugmentor$ consists of a base I2V DiT model $\basemodel$ and a sparse control branch (illustrated in \cref{fig:method_sparse}).
We follow the I2V encoding pipeline from \cref{sec:prelim}, but with several adaptations.
Firstly, we adapt the original video caption into a motion-only prompt (no appearance cues) so that the generated video adheres to our mask color palette.
Secondly, besides the first-frame HOI mask $\inputhoimask$, we feed the original image $\inputimg$ as an additional condition to provide appearance context.
Consequently, the patch embedder~$\patchembaug$ is finetuned to accommodate the increased channel dimensions. 

The sparse control branch in \cref{fig:method_sparse} takes as input the blurred trajectory rendering $\trajvideoblur$ and the blurred visibility map $\vismaskblur$.
The trajectory rendering $\trajvideoblur$ is first encoded by the VAE and then processed by the trajectory extractor $\trajextractor$ to produce trajectory features $\trajfeature$.
The visibility map $\vismaskblur$ serves as an auxiliary reliability cue and is directly downsampled to the spatial resolution of $\trajfeature$, avoiding unnecessary VAE compression.
To inject these motion cues into the diffusion backbone, the augmentor fuser $\augmentorfuser$ predicts FiLM scale and shift parameters~\cite{zhang2025tora,perez2018film}:
\begin{equation}
\augscale,\augshift = \augmentorfuser(\trajfeature).
\end{equation}
$\augscale$ and $\augshift$ modulate each DiT block and determine how strongly the sparse trajectory information influences generation (see \supmat for FiLM conditioning preliminaries).
Trajectory features can be unreliable due to occlusion, out-of-frame motion, tracker drift, or reduced precision from Gaussian blurring, which would cause FiLM modulation to rely on degraded cues.
We introduce a visibility-aware gating mechanism that adaptively adjusts the influence of the trajectory features.
The visibility map $\vismaskblur$ is embedded using the visibility embedder $\visemb$ to produce visibility features and forms a learnable gate $\gatevis$:
\begin{equation}
\begin{aligned}
\visibilityfeature &= \visemb(\vismaskblur), \\
\gatevis &= \vismaskblur + (1 - \vismaskblur) \odot \sigma(\visibilityfeature),
\end{aligned}
\end{equation}
where $\odot$ denotes element-wise multiplication and $\sigma$ is the sigmoid function.
$\gatevis \in [0,1]$ acts as a confidence score that modulates the FiLM parameters:
\begin{equation}
\augscalegated = \augscale \odot \gatevis, \qquad
\augshiftgated = \augshift \odot \gatevis.
\end{equation}
Finally, the gated FiLM parameters modulate the DiT tokens before the attention layer in each block (see \supmat \cref{fig:inside_dit}).
\begin{equation}
\dittokenmodulated = \dittoken + \dittoken \odot \augscalegated + \augshiftgated.
\end{equation}
This visibility-aware fusion allows the network to follow the intended motion while maintaining robustness to noise.
 \begin{figure}[!hbt]
 \centering
	\includegraphics[width=1\linewidth]{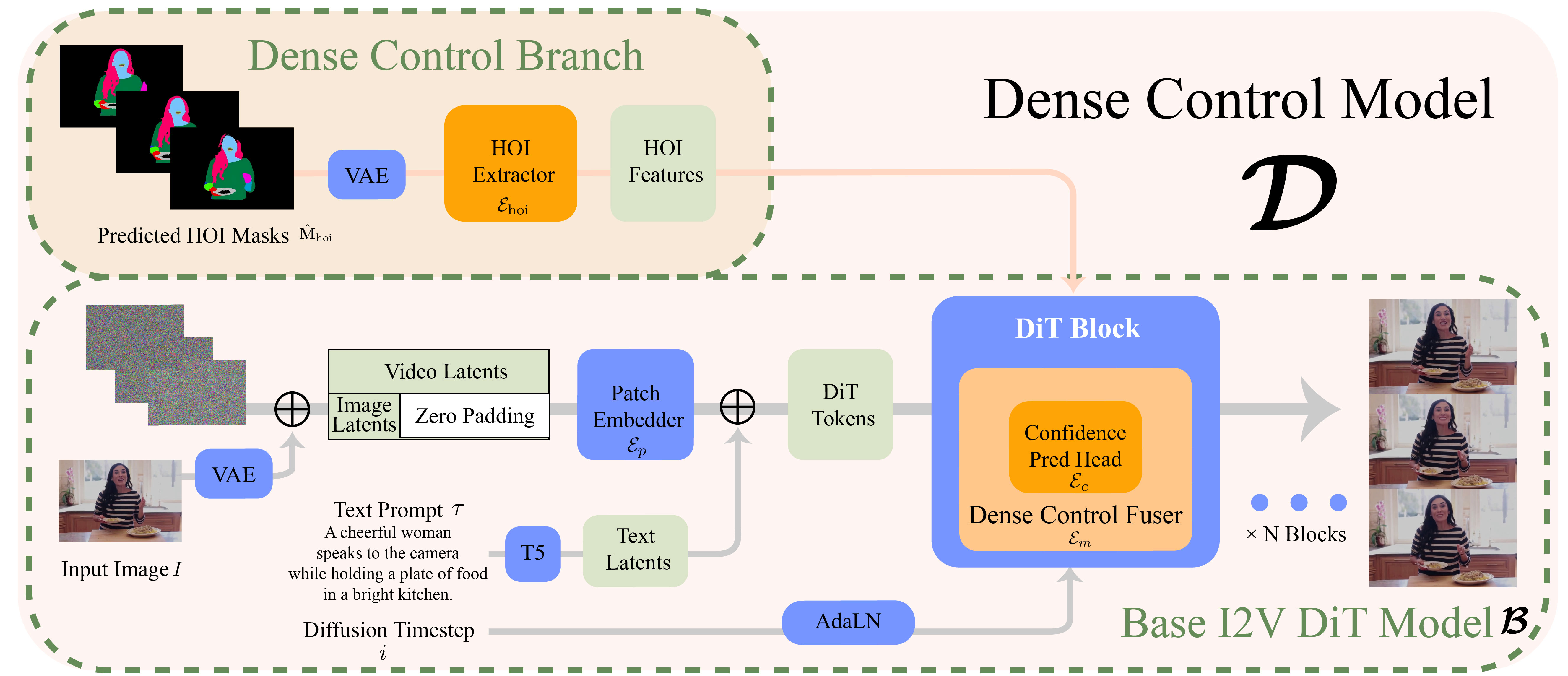}
        \caption
	{
        \textbf{The dense control model} $\densemodel$ conditions on HOI masks. 
The masks are encoded by a HOI extractor and fused with transformer latents in the dense control fuser, which also includes a confidence prediction head to modulate reliance on the control signal. 
The final output is an HOI video that follows the densified motion cues.
Orange modules denote learnable components; blue modules are frozen. (Best viewed with zoom)} 
	\label{fig:method_dense}
\end{figure}

\subsection{Dense Control Model}
\label{sec:dense_model}
Given the predicted HOI mask sequence $\hoimaskseqpred$ from the augmentor
$\trajaugmentor$, the second stage of \methodname{} uses these densified,
instance-aware motion cues to generate the final HOI video $\outputvideo$
(\cref{fig:method_dense}).  
The purpose of this stage is to translate the geometry-based intermediate
representation into realistic appearance and interaction dynamics.  
Because $\hoimaskseqpred$ describes how individual body parts and the object
move, deform, and contact over time without being entangled with textures or lighting, it provides a structured signal that simplifies the generation task and enables more coherent human-object interactions.
The dense control model $\densemodel$ adopts the same diffusion backbone and control branch design as the augmentor $\trajaugmentor$, but differs in two key respects:

1) The diffusion backbone $\basemodel$ now operates in a standard
image-to-video (I2V) setting, conditioned only on the text prompt $\textprompt$
and input image $\inputimg$, without an additional context frame as in $\trajaugmentor$.  
Consequently, the patch embedder $\patchemb$ remains frozen with no channel
expansion, and the network outputs a full HOI video rather than a mask
sequence.

2) The dense control branch receives the predicted HOI mask sequence
$\hoimaskseqpred$ and processes it through a HOI extractor $\hoiextractor$ to
obtain HOI features $\hoifeature$, which are fused with the DiT tokens within
each transformer block.

\methodname{} further introduces a confidence-based fusion mechanism through a dense control fuser $\maskfuser$.  
This is because human segmentation masks used during training (e.g., from \cite{khirodkar2024sapiens}) may flicker, be misaligned, or be incomplete; 
such noise may be propagated to the predicted $\hoimaskseqpred$, hence over-reliance on such noisy masks is undesirable. 
We therefore add a confidence head $\confidencehead$ that predicts a scalar $\confidence \in [0,1]$ to modulate HOI feature influence:
\begin{equation}
\confidence = \sigma\left(\confidencehead(\hoifeature)\right), \qquad
\hoifeaturegated = \confidence \cdot \hoifeature,
\end{equation}
where $\sigma$ is the sigmoid function.
The gated HOI features $\hoifeaturegated$ downweights unreliable conditioning, improving robustness to prediction errors from $\trajaugmentor$ as well as noise in the ground-truth training masks.
$\hoifeaturegated$ is then passed to the FiLM layers to modulate the transformer tokens $\dittoken$.

In summary, $\densemodel$ uses the densified HOI masks as structured motion guidance to produce appearance-level interaction dynamics, forming the final stage of our sparse-to-dense controllable HOI generation framework and enabling coherent, realistic, and robust HOI synthesis.

\section{Implementation Details}
\label{sec:implement}
The trainable components in $\densemodel$ and $\trajaugmentor$ include $\hoiextractor$, $\trajextractor$, 
$\augmentorfuser$, $\maskfuser$,  $\visemb$ and $\confidencehead$, with each component independently instantiated.
$\hoiextractor$ and $\trajextractor$ have the same architecture, but they do not share weights.
The same applies to the fusers $\augmentorfuser$ and $\maskfuser$.
Our HOI extractor $\hoiextractor$ and trajectory extractor $\trajextractor$ follow the architecture of~\cite{zhang2025tora}, using separate ResNet blocks per DiT layer. 
To improve training stability under small batch sizes, we replace all BatchNorm~\cite{ioffe2015batch} layers with LayerNorm~\cite{ba2016layer}. 
The augmentor fuser $\augmentorfuser$, dense control fuser $\maskfuser$, and visibility embedder $\visemb$ use spatial and temporal convolution layers, while the confidence head $\confidencehead$ consists of two 3D convolution layers.
The diffusion backbone $\basemodel$ remains frozen throughout training.
For training the dense control model $\densemodel$, we apply a dropout strategy to improve robustness to missing or incomplete motion cues: in 50\% of cases we provide both human and object masks, in 20\% only human masks, in 20\% only object masks, and in 10\% we drop all conditioning.
For augmentor training, we randomly sample the number of trajectories as 
$\numtrajectories \in \{1, \dots, 8\}$  
for both human and object points.
We fine-tune on a 35k subset of HOIGen-1M~\cite{liu2025hoigen} for one epoch using 4 A100, H100, or H200 GPUs in parallel.
The training takes 2 days on H200.
For results on BEHAVE~\cite{bhatnagar22behave}, we further fine-tune for four epochs on a 4k-video subset of the dataset. 
The effective batch size is 8 with gradient accumulation, and the learning rate is $5\!\times\!10^{-5}$.
To obtain the blurred visibility maps $\vismaskblur$ and trajectory renderings $\trajvideoblur$, we apply a Gaussian smoothing kernel with radius $\kernelradius = 99$ and standard deviation $\gaussiansmooth = 10$.
During inference, generating a 49×480×720 video takes $\sim$10 minutes on a single H100 GPU.

\section{Experiments}
\label{sec:experiments}
\begin{table*}[hbt]
\centering
\resizebox{0.8\linewidth}{!}
{
\begin{tabular}{llrrrrrrrrrr}
    \toprule
    & \multirow{2}{*}{Method} & \multirow{2}{*}{FVD$\downarrow$} & \multirow{2}{*}{TE$\downarrow$} & \multirow{2}{*}{CA$\uparrow$} & \multirow{2}{*}{CLIPSIM$\uparrow$} & \multicolumn{6}{c}{VBench} \\
    \cmidrule(lr){7-12}
    &  &  &  &  &  & SC$\uparrow$ & BC$\uparrow$ & DD$\uparrow$ & MS$\uparrow$ & AQ$\uparrow$ & IQ$\uparrow$ \\
    \midrule
    & GT                                              & N/A & N/A & N/A & 0.3040 & 0.96 & 0.95 & 0.06 & 1.00 & 0.48 & 0.75 \\
    \midrule
    \multirow{4}{*}{Comparison} & TORA~\cite{zhang2025tora}               &1359 & 12.95      & 0.775 & 0.2918 & 0.90 & \second{0.93} & 0.56 & \second{0.98} & 0.49 & 0.63 \\
     & TORA-Finetuned                                     &\second{1078} & \best{9.87}       & \best{0.828} & 0.3033 & \second{0.92} & \second{0.93} & \best{0.74} & \second{0.98} & \best{0.51} & 0.64 \\
     & Go-with-the-flow~\cite{burgert2025go} &1312 & 12.85      & 0.789 & \second{0.3035} & \best{0.93} & \best{0.94} & 0.42 & \best{0.99} & \second{0.50} & \second{0.67} \\
     & \ours{\textbf{\methodname{}}}                                               &\best{915}  & \second{10.64}      & \second{0.827} & \best{0.3036} & \best{0.93} & \best{0.94} & \second{0.58} & \best{0.99} & \best{0.51} & \best{0.68} \\ 
    \midrule
    \multirow{9}{*}{Ablation} & Dense flow (upper bound)                          &697 & 5.21        & 0.931 &0.3043 & 0.92 & 0.94 & 0.72 & 0.98 & 0.50 & 0.66 \\
     & Dense mask (upper bound)                          &708 & 6.12        &0.908  &0.3044 &0.92 & 0.94 & 0.70 & 0.98 & 0.51 & 0.67 \\
     & Dense HOI  (upper bound)                          &675 & 6.44        &0.917  &0.3038 &0.92 & 0.94 & 0.74 & 0.98 & 0.51 & 0.67 \\
     \cmidrule(lr){2-12}
     & Sparse HOI                                        &993 & 9.18        &0.860  &0.3052 & 0.93 & 0.94 & 0.76 & 0.98 & 0.51 & 0.67 \\
     & Augmented flow                                    &1055&14.93       &0.784  &0.3047 &0.92 & 0.93 & 0.68 & 0.98 & 0.50 & 0.66 \\    
     & Augmented mask                                    &971 &11.49       &0.822  &0.3037 &0.92 & 0.94 & 0.56 & 0.98 & 0.51 & 0.67  \\ 
     \cmidrule(lr){2-12}
     & \methodname{}  w/o first image                             &980 & 9.28        &0.782 &0.3032 &0.92 & 0.94 & 0.56 & 0.99 & 0.51 & 0.68 \\
     &  \methodname{} w/o conf                                    &1004 & 9.11        & 0.807 &0.3031 &0.93 & 0.94 & 0.58 & 0.98 & 0.51 & 0.68 \\
     & \methodname{}  w/o vis, w/o conf                           &941 &10.36       & 0.833 &0.3050 &0.92 & 0.94 & 0.64 & 0.98 & 0.51 & 0.68 \\
    \bottomrule
\end{tabular}
}
\caption{
\textbf{Baseline Comparisons and Ablation Studies on HOIGen-1M.} The top section compares our method against TORA, TORA-Finetuned, and Go-with-the-Flow. We also provide the GT numbers as a reference in the first row. We use red to highlight the top-performing method, and orange denotes the second best.
Our method beats the baselines in terms of FVD, CLIPSIM, SC, BC, MS, AQ and IQ, indicating an improvement in video quality.
The second section reports upper-bound performance using ground-truth dense signals from RAFT~\cite{teed2020raft}, SAM~\cite{ravi2024sam2segmentimages}, and SAPIEN~\cite{khirodkar2024sapiens}, instead of predicted ones. The third section evaluates sparse HOI control, and shows that augmenting either masks or flow consistently improves over baselines in the top section, although flow yields lower perceptual quality (higher FVD). The final section ablates three design choices: removing the first-frame appearance input from the augmentor, removing the confidence head, and removing visibility gating together with confidence head. 
} 
\label{tab:quantitative_baselines_hoigen}	
\end{table*}

%
\begin{table}[hbt]
\centering
\resizebox{\linewidth}{!}
{
\begin{tabular}{lrrrrrrrrrrr}
    \toprule
    \multirow{2}{*}{Method} & \multirow{2}{*}{FVD$\downarrow$} & \multirow{2}{*}{TE$\downarrow$} & \multirow{2}{*}{CA$\uparrow$} & \multirow{2}{*}{CLIPSIM$\uparrow$} & \multicolumn{6}{c}{VBench} \\
    \cmidrule(lr){6-11}
     &  &  &  &  & SC$\uparrow$ & BC$\uparrow$ & DD$\uparrow$ & MS$\uparrow$ & AQ$\uparrow$ & IQ$\uparrow$ \\
    \midrule
      GT                                     & N/A & N/A         & N/A & 0.3138 & 0.96 & 0.94 & 0.00 & 1.00 & 0.53 & 0.72 \\
     \midrule
    TORA~\cite{zhang2025tora}                &675  & 7.07       & 0.673 & 0.3069 &0.94 & 0.94 & \best{0.06} & \second{0.99} & \best{0.51} & \best{0.69}\\
    TORA-Finetuned                           &\second{532}  & \second{6.68}       & \best{0.803} & \best{0.3083} &\second{0.95} & \second{0.95} & \second{0.02} & \second{0.99} & \best{0.51} & \best{0.69}\\
    Go-with-the-flow~\cite{burgert2025go}    &737  & 8.51       & 0.686 & \second{0.3078} &\best{0.96} & \best{0.96} & 0.00  & \best{1.00}   & \best{0.51}& \second{0.66} \\
    \ours{\textbf{\methodname{}}}                                     &\best{434} & \best{5.34}       & \second{0.721} & 0.3061 &\best{0.96} & \second{0.95} & 0.00 & \best{1.00} & \best{0.51} & \best{0.69} \\ 
    \bottomrule
\end{tabular}
}
\vspace{-3mm}
\caption{
\textbf{Quantitative Comparisons on BEHAVE dataset.} Our method leads on FVD and TE and matches the top scores on SC, MS, AQ, and IQ. We use red to highlight the top-performing method, and orange denotes the second best.
} 
\label{tab:quantitative_baselines_behave}	
\end{table}

%
\begin{table}[hbt]
\centering
\vspace{-5mm}
\resizebox{\linewidth}{!}
{
\begin{tabular}{lrrrrrrrrrrr}
    \toprule
    \multirow{2}{*}{Method}                                             & \multirow{2}{*}{FVD$\downarrow$} & \multirow{2}{*}{TE$\downarrow$} & \multirow{2}{*}{CA$\uparrow$} & \multirow{2}{*}{CLIPSIM$\uparrow$} & \multicolumn{6}{c}{VBench} \\
    \cmidrule(lr){6-11}
     &  &  &  &  & SC$\uparrow$ & BC$\uparrow$ & DD$\uparrow$ & MS$\uparrow$ & AQ$\uparrow$ & IQ$\uparrow$ \\
    \midrule
      GT                                               & N/A & N/A & N/A & 0.2850 &0.96 & 0.94 & 0.28 & 0.99 & 0.50 & 0.72  \\
     \midrule
    TORA~\cite{zhang2025tora}              & 726 & 10.26       & \best{0.746} &0.2813 &0.93 & \second{0.92} & \second{0.30} & \best{0.99} & \second{0.48} & \best{0.70}     \\
    TORA-Finetuned                                     & \second{606} & \second{9.43}        & 0.716 &\second{0.2842} &\second{0.94} & \best{0.93} & \best{0.36} & \best{0.99} & \second{0.48} & \second{0.69}  \\
    Go-with-the-flow~\cite{burgert2025go}  & 780 & 11.90        & 0.707 &0.2806 &\best{0.95} & \best{0.93} & 0.22 & \best{0.99} & \best{0.49} & \second{0.69}  \\
    \ours{\textbf{\methodname{}}}                                                   & \best{484} & \best{6.79}        & \second{0.721} &\best{0.2854} &\best{0.95} & \best{0.93} & \second{0.30} & \best{0.99} & \second{0.48} & \best{0.70}  \\
        \bottomrule
\end{tabular}
}
\vspace{-3mm}
\caption{
\textbf{Quantitative Comparisons on in-the-wild dataset.} Our method is best on FVD, TE, CLIPSIM, SC, BC, and IQ. We use red to highlight the top-performing method, and orange denotes the second best.
} 
\label{tab:quantitative_baselines_wild}	
\end{table}

 \begin{figure*}[!hbt]
 \centering
	\includegraphics[width=1\linewidth]{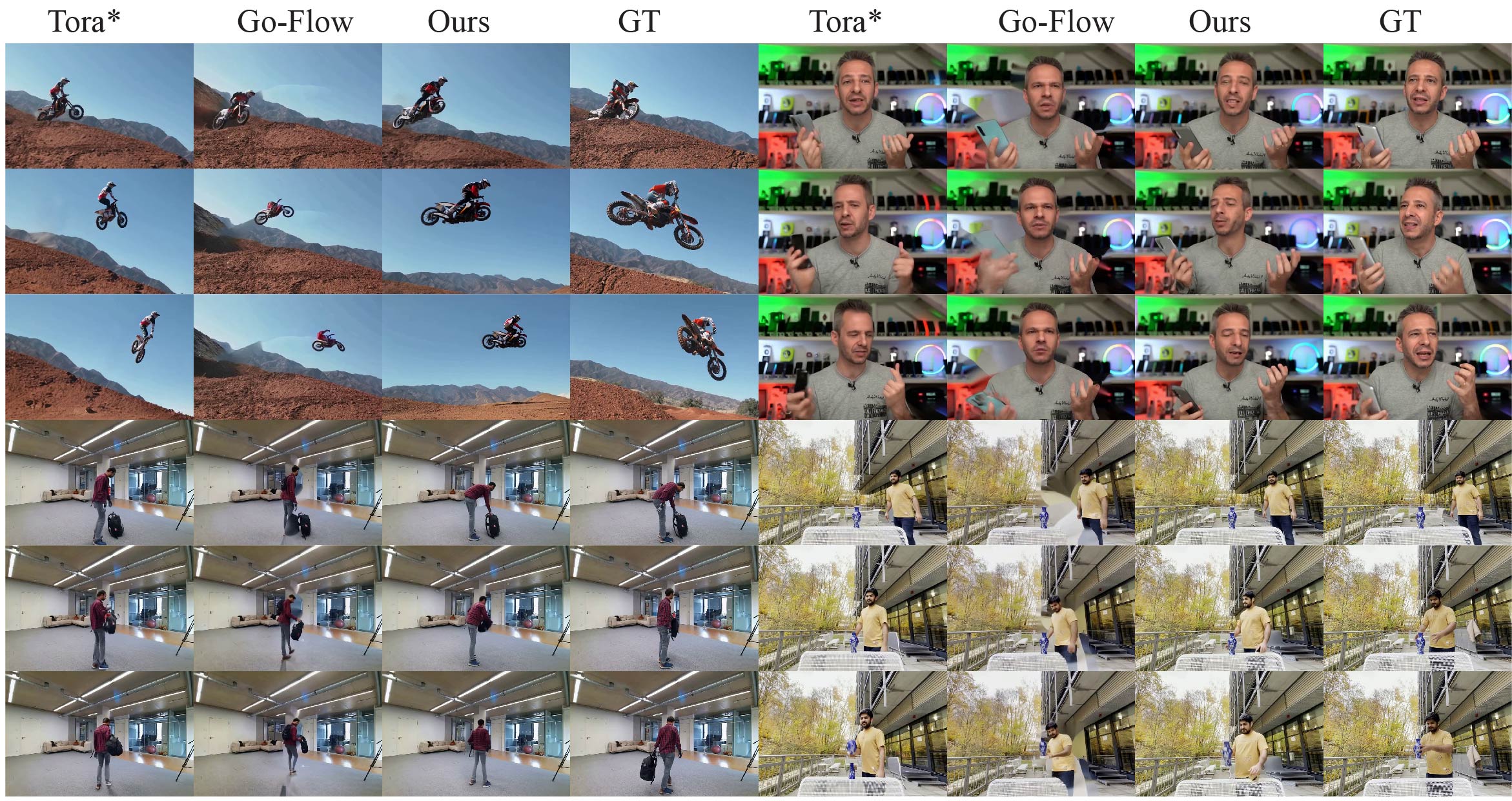}
        \caption
	{
Qualitative comparisons of TORA-finetuned (TORA*), Go-with-the-Flow (Go-Flow), and our method alongside ground-truth videos. Our approach achieves higher interaction fidelity and visual quality across diverse HOI scenarios.
        }
	\label{fig:qualitative_comparison}
\end{figure*}

 \begin{figure}[!hbt]
 \vspace{-5mm}
 \centering
	\includegraphics[width=1\linewidth]{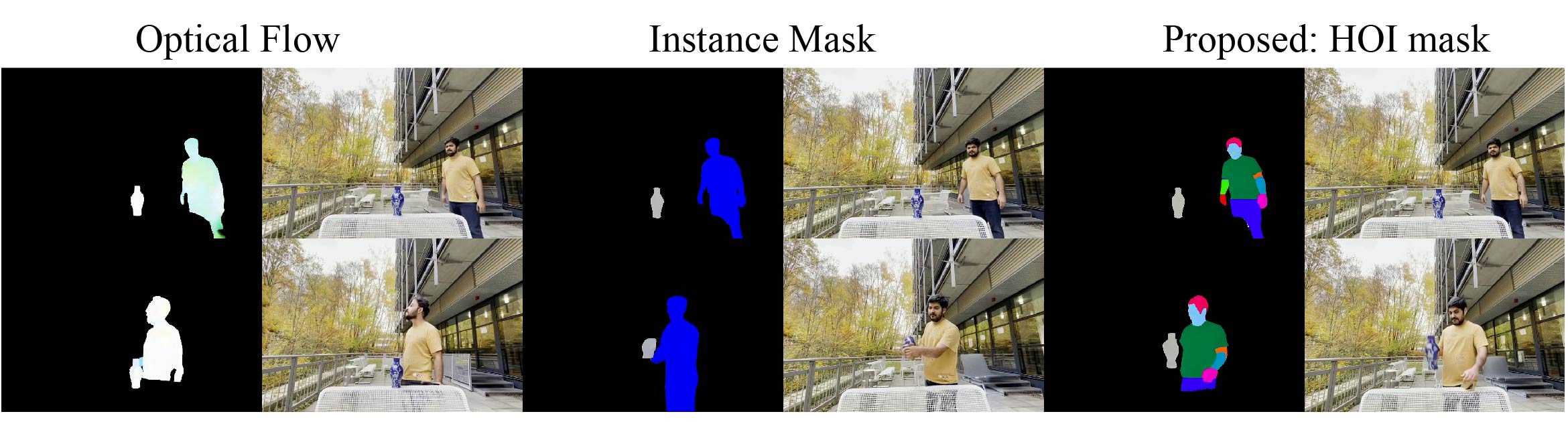}
        \caption
	{
\textbf{Qualitative ablation of different motion representations.}
We compare augmentors trained on foreground optical flow, instance masks, and our proposed HOI masks. Flow-based conditioning lacks interaction semantics and fails to capture the grasp in this example. Instance-mask conditioning predicts the interaction but does not preserve object identity. Our HOI mask representation provides richer interaction semantics and leads to higher-quality video generation.}
\vspace{-5mm}
	\label{fig:ablation}
\end{figure}

\subsection{Evaluation Datasets}
We evaluate on three datasets, each containing 50 videos with diverse objects and interaction behaviors.
First, we sample 50 videos from the hold-out split of HOIGen-1M~\cite{liu2025hoigen}.
Second, we use 50 videos from the BEHAVE dataset~\cite{bhatnagar22behave}, which includes challenging cases where the human is not initially in contact with the object, requiring target-aware navigation before interaction.
Finally, we curate 50 in-the-wild videos with varied subjects and object categories to assess generalization in real-world scenarios.

\subsection{Baselines}
We compare against recent DiT-based trajectory-controlled video generation methods that operate under comparable input assumptions: (1) \textbf{TORA}~\cite{zhang2025tora}: a trajectory-conditioned video generation model.
(2) \textbf{Go-with-the-Flow}~\cite{burgert2025go}: a diffusion conditioning method based on noise warping.
We adapt it to our setting by providing first-frame masks and rigidly transforming the mask using the input trajectory.
We initialize our model from TORA’s publicly released checkpoint and fine-tune it with our newly introduced modules and motion representations.
For fairness, we report results from both the original TORA model and our re-trained TORA variant fine-tuned on HOIGen-1M.

\subsection{Evaluation Metrics}
Following prior works in controllable video generation~\cite{huang2023vbench, kim2025target, li2025genhsi, zhang2025tora}, we adopt a comprehensive suite of metrics to evaluate different aspects of video quality. 
For general video generation, we report VBench metrics~\cite{huang2023vbench}, including subject consistency (SC), background consistency (BC), dynamic degree (DD), motion smoothness (MS), aesthetics quality (AQ), and imaging quality (IQ). 
We additionally measure overall video fidelity using FVD~\cite{unterthiner2018towards} and text-video alignment using CLIPSIM~\cite{radford2021learning, wang2024motionctrl}. 
To evaluate controllable HOI synthesis, we introduce two domain-specific metrics, i.e 
\textbf{Contact Accuracy (CA):} the per-frame hand-object contact correctness relative to ground truth using a contact detector~\cite{narasimhaswamy2020detecting}, and the
\textbf{Trajectory Error (TE):} 
the Euclidean distance between the ground-truth trajectory and the generated video obtained via CoTracker3~\cite{karaev2024cotracker3}.
\subsection{Qualitative Comparisons}
We provide qualitative comparisons in \cref{fig:qualitative_comparison}.
Go-with-the-Flow produces background blur near motion regions, while TORA* suffers from low interaction fidelity, including blurry hands and implausible object transformation. Our method achieves more coherent and realistic interactions. We refer the reader to the supplementary video for demonstrations of trajectory control accuracy.
\subsection{Quantitative Comparisons}
Among all metrics in \cref{tab:quantitative_baselines_hoigen}, \cref{tab:quantitative_baselines_behave}, and \cref{tab:quantitative_baselines_wild}, we find FVD, TE, and CA to be the most informative for evaluating HOI quality. In contrast, VBench and CLIPSIM show smaller differences between methods.
Across three datasets which span a wide range of HOI scenarios, our method consistently outperforms baselines on FVD and, in most cases, on TE and CA, indicating improved video quality and more realistic interactions.
We further include a user study in \supmat, which corroborates these findings.
Interestingly, we also find that a simple strategy of overlaying sparse trajectories on the segmentation masks for dense model training can further boost the performance (\supmat \cref{sec:traj_aug}).

\subsection{Ablations}
\paragraph{Ablation setup.} We conduct ablation studies to understand the contribution of each component in our framework. Specifically, we investigate:
\begin{enumerate}
    \item \textbf{Sparse vs. dense control.} What is the performance gap between sparse and dense conditioning, and to what extent does our augmentor bridge this gap?
    \item \textbf{Motion representation.} Which control signal is most effective for HOI video synthesis: foreground optical flow, instance-based coarse human-object masks, or our proposed part-level HOI masks?
    \item \textbf{First-frame image context.} Does supplying the first input image (in addition to the HOI mask) to the augmentor improve trajectory densification and downstream HOI synthesis quality?
    \item \textbf{Visibility and confidence gating.} How important are the visibility gate and confidence gate for the network in terms of handling noisy motion signal?
\end{enumerate}
\vspace{-5mm}
\paragraph{Findings.}
We report ablation results in \cref{tab:quantitative_baselines_hoigen} and summarize the key findings as follows:

\begin{enumerate}
    \item \textbf{Sparse vs.\ dense control.}  
    We observe a performance gap between sparse trajectory conditioning (Sparse HOI) and dense HOI masks (upper bound), highlighting the importance of densifying sparse motion cues to obtain more instance and interaction-aware guidance.
    \item \textbf{Choice of motion representation.}  
    Using the augmentor to densify any representation improves performance, though with different trade-offs.  
    Foreground optical flow offers the finest granularity but is highly sensitive to camera motion, and its color encoding reflects only motion direction/magnitude rather than HOI semantics. As a result, flow augmentors are more difficult to train, and errors propagate strongly to the dense model.  
    Instance masks also improve performance but lack the finer semantic detail captured by our HOI-aware masks.  
    Our HOI representation, tailored for interaction reasoning, achieves the best FVD, TE, and CA. Its limitation is that SAPIEN-derived training masks~\cite{khirodkar2024sapiens} are less smooth than RAFT optical flow~\cite{teed2020raft} or SAM-based masks~\cite{ren2024grounded}, which can affect prediction quality.

    \item \textbf{Image context for the augmentor} ($\inputimg$)  
    is important for grounding the interaction context. Removing it leads to a noticeable drop in CA.

    \item \textbf{Visibility and confidence gating.}  
    Both gating mechanisms improve robustness to noisy or imperfect signals, though neither is strictly critical for model performance. Even without visibility input, our method remains competitive and outperforms the baselines, while full gating provides additional stability.
\end{enumerate}

\section{Conclusion}
We presented \methodname, a two-stage augment-and-control framework for controllable HOI video generation. 
Our key insight is that introducing a learnable intermediate motion representation, i.e densified HOI masks, enables the model to better capture the HOI spatiotemporal dynamics. 
This design effectively bridges the gap between user-friendly sparse trajectories and the rich motion cues required for high-quality synthesis, making our approach suitable for downstream creative and animation workflows.
Our method achieves state-of-the-art motion naturalness and interaction fidelity across diverse HOI scenarios.
\vspace{-5mm}
\paragraph{Limitation and Future Work}
While effective, our results remain bounded by limitations inherited from the underlying video diffusion backbone, including imperfect identity preservation, occasional physical implausibility, and limited 3D awareness. 
Additionally, inaccuracies in the auxiliary segmentation~\cite{khirodkar2024sapiens, ren2024grounded} and tracking~\cite{karaev2024cotracker3} models used during training to derive control signals may introduce errors that propagate to the augmentor’s predictions.
Representative failure cases are provided in the supplementary video.
Nevertheless, our sparse-to-dense control design and HOI motion representation are model-agnostic and can be readily integrated into future, stronger video generators and perception models, promising further improvements in HOI fidelity and controllability.
\newline
\textbf{Acknowledgements:} The authors would like to thank Amin Parchami for helpful advice on cluster usage.
{
    \small
    \nocite{*}
    \bibliographystyle{ieeenat_fullname}
    \bibliography{main}
}

\clearpage
\maketitlesupplementary
The supplementary material is organized as follows.
We begin with FiLM preliminaries and a more detailed diagram of the fuser placement inside the DiT block in \cref{sec:inside_dit}, followed by a perceptual user study in \cref{sec:inside_dit}.
We provide additional HOI mask generation quality analysis in \cref{sec:mask_iou}.
We then detail the comparison to a UNet-based model in \cref{sec:comparison_motioni2v}.
Next, we visualize the HOI mask color palette in \cref{sec:color_palette}, and provide the prompt template in \cref{sec:augmentor_prompt}, which is used to process the text prompt for training the augmentor.
\section{FiLM preliminaries}
Feature-wise Linear Modulation (FiLM)~\cite{perez2018film} augments a network backbone with a learned feature-wise affine operator that conditions on an auxiliary signal 
without altering the backbone topology. 
Let $\boldsymbol{x}\in\mathbb{R}^{N\times D}$ denote an intermediate feature matrix (e.g., a sequence of $N$ tokens with feature dimension $D$) and $\boldsymbol{z}\in\mathbb{R}^{d}$ a conditioning embedding (e.g., motion features in our case). 
FiLM realizes a mapping $\psi:\mathbb{R}^{d}\rightarrow\mathbb{R}^{2D}$ that yields feature-wise parameters $(\boldsymbol{\gamma},\boldsymbol{\beta})=\psi(\boldsymbol{z})$, and applies them as
\begin{equation}
    \boldsymbol{x}' = \boldsymbol{x}\odot\boldsymbol{\gamma} + \boldsymbol{\beta},
\end{equation}
FiLM implements a conditional reparameterization that aligns feature responses with the semantics encoded in the conditioning stream, 
leading to tight cross-modal coupling.
In addition, it is also lightweight and architecture-agnostic, making it a popular choice for conditioning large pretrained backbones in a parameter-efficient manner.

In \methodname{}, FiLM modulation is applied in both stages with stage-specific motion conditioning.
Specifically, in the trajectory augmentor $\trajaugmentor$, the fuser $\augmentorfuser$ predicts motion-dependent scale/shift fields from $\trajfeature$, applies visibility gating to obtain $\augscalegated,\augshiftgated$, and injects them into visual tokens via normalized residual modulation.
In the dense control model $\densemodel$, $\maskfuser$ and the confidence branch analogously produce gated HOI conditioning from $\hoifeaturegated$ for the same pre-attention modulation pathway.
Therefore, a shared FiLM-style design is used to inject sparse or dense HOI motion cues through lightweight conditional affine modulation.
\section{Inside the DiT Block}
\label{sec:inside_dit}
\begin{figure}[!t]
	\centering
	\includegraphics[width=0.9\columnwidth]{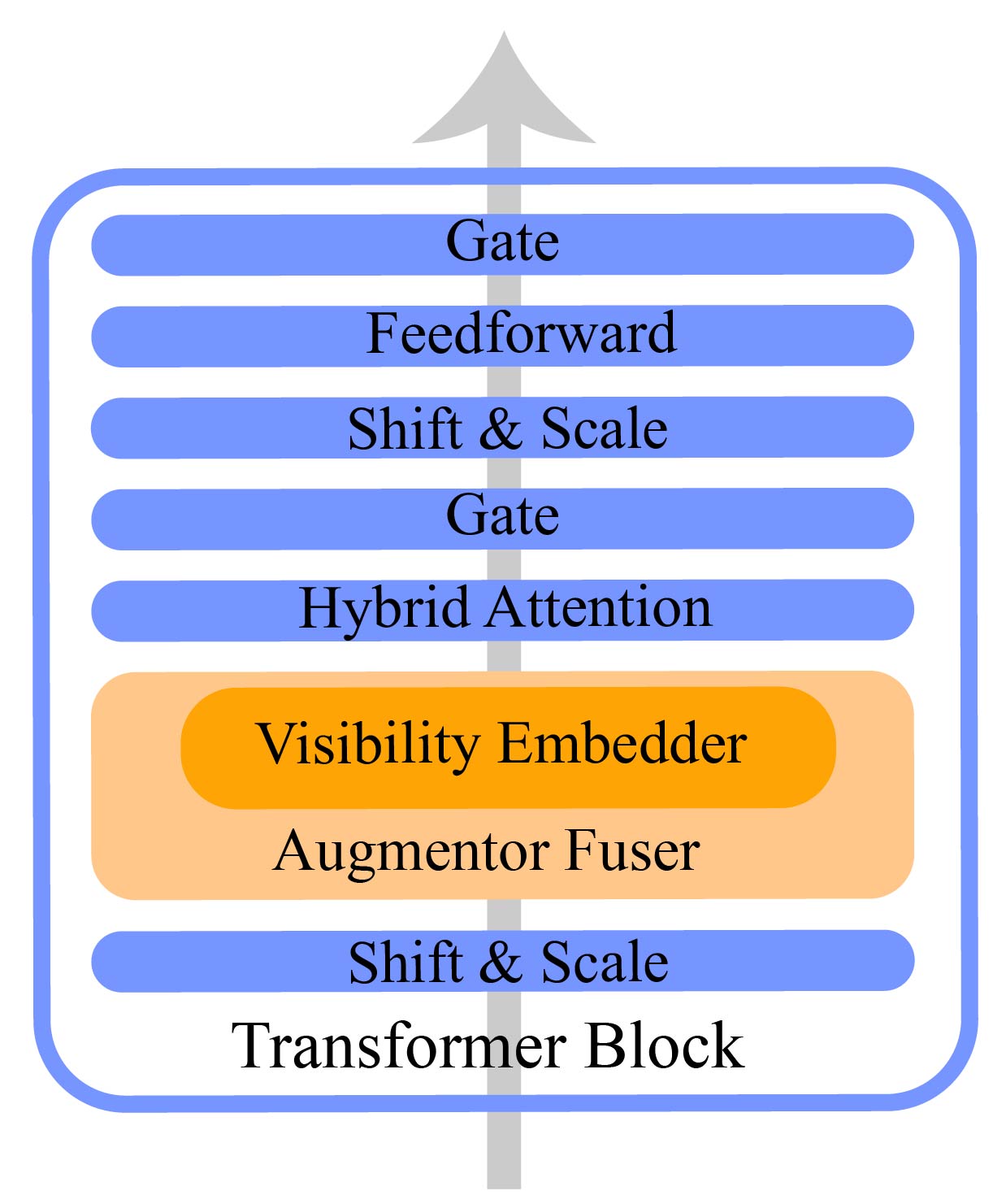}
	\caption{
        The fuser is inserted between the AdaLN scale-and-shift and the hybrid attention. The same placement is used in the dense model. The fuser is fine-tuned, while all other layers remain frozen.
        }
	\label{fig:inside_dit}
\end{figure}

We provide a detailed diagram in~\cref{fig:inside_dit} illustrating where the fuser is inserted into the DiT block for the augmentor. The DiT tokens are first modulated by the AdaLN scale and shift operation, with fusion occurring before the attention layer.

\section{Perceptual User Study}
\label{sec:user_study}
\begin{figure}[!t]
	\centering
	\includegraphics[width=\columnwidth]{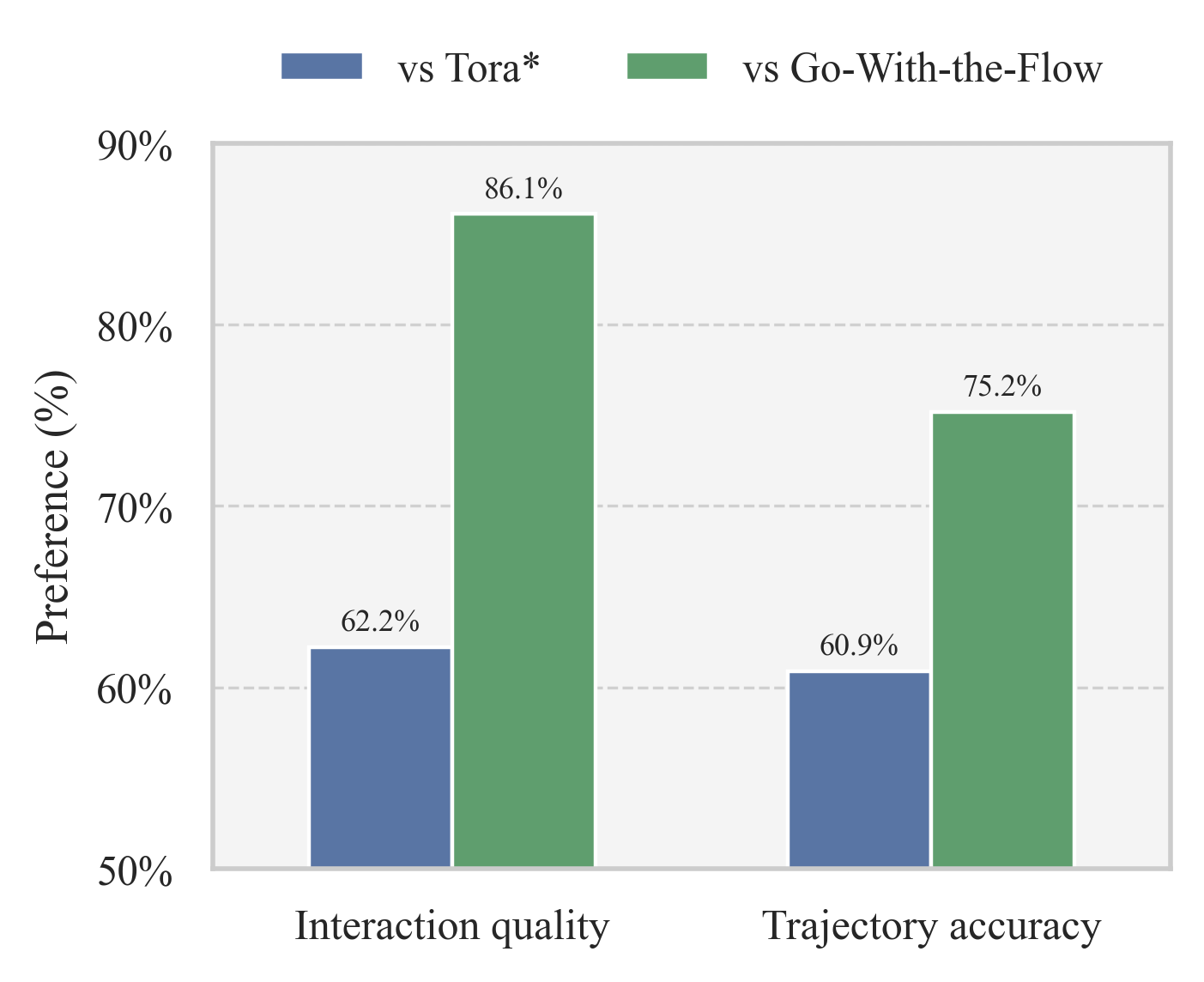}
	\caption{
        \textbf{User Study Results.}
		VHOI is preferred over Tora* (finetuned) and Go-With-the-Flow in terms of both human-object interaction quality and trajectory adherence.
        }
	\label{fig:user_study}
\end{figure}

Despite the efforts in benchmarking video generation quality, existing quantitative metrics still miss nuances of interaction realism.
To complement them, we conduct a user study with 46 participants.
The survey consists of 20 questions in total, and is split into two groups of 10 questions.
Each comparison presents our video alongside either TORA* or Go-With-the-Flow in random order.
Group 1 asks: \textit{Which video exhibits more natural human-object interaction?}.
Group 2 shows videos with overlaid trajectories, and asks \textit{Which video follows the trajectories more accurately?}
Aggregated preferences are shown in~\cref{fig:user_study}.
Participants favor our method in 62.2\% of interaction comparisons vs TORA* and 86.1\% vs. Go-With-the-Flow, and in 60.9\% and 75.2\% of trajectory comparisons, respectively.
Two-sided binomial $z$-tests with $\alpha=0.05$ confirm that all improvements are statistically significant.
Qualitatively, Go-With-the-Flow often produces rigid shifts and incoherent motion, whereas TORA* is competitive yet frequently lacks precise instance awareness during contact-rich interactions.

\section{Comparison with MotionI2V}
\label{sec:comparison_motioni2v}
We additionally compare our method against MotionI2V~\cite{shi2024motion}, a two-stage controllable video generation framework based on a UNet backbone with optical-flow based densification.
Because MotionI2V operates at a lower resolution ($16 \times 512 \times 312$) than ours ($49 \times 720 \times 490$), we downsample our generated videos and compute the trajectory error (TE), and contact accuracy (CA) at their native resolution for a fair comparison, while the rest of the metrics are reported with the respective native resolution. 
Quantitative evaluations are presented in \cref{tab:quantitative_motioni2v}. 
The qualitative results of MotionI2V exhibit restricted motion and reduced coherence; further qualitative comparisons are included in the supplementary video.
%
\begin{table}[hbt]
\centering
\resizebox{\linewidth}{!}
{
\begin{tabular}{lrrrrrrrrrrr}
    \toprule
    \multirow{2}{*}{Method}                                             & \multirow{2}{*}{FVD$\downarrow$} & \multirow{2}{*}{TE$\downarrow$} & \multirow{2}{*}{CA$\uparrow$} & \multirow{2}{*}{CLIPSIM$\uparrow$} & \multicolumn{6}{c}{VBench} \\
    \cmidrule(lr){6-11}
     &  &  &  &  & SC$\uparrow$ & BC$\uparrow$ & DD$\uparrow$ & MS$\uparrow$ & AQ$\uparrow$ & IQ$\uparrow$ \\
    \midrule
    MotionI2V~\cite{shi2024motion}                     &1629 & 19.06& 0.745 & \best{0.3054} &\best{0.95} & \best{0.96} & 0.26 & \best{0.99} & 0.47 & 0.65   \\
    VHOI (Ours)                                 &\best{915}  & \best{14.82}    & \best{0.830} & 0.3036& 0.93 & 0.94 & \best{0.58} & \best{0.99} & \best{0.51} & \best{0.68} \\   
        \bottomrule
\end{tabular}
}
\caption{
\textbf{Quantitative Comparison with MotionI2V.} As evidenced by the significantly higher FVD and CA and lower TE, MotionI2V produces less realistic interactions compared with \methodname{}.
} 
\label{tab:quantitative_motioni2v}	
\end{table}

\section{HOI Mask Quality Evaluation}
\label{sec:mask_iou}
We evaluate how well the generated HOI masks match the ground truth on the HOIGen-1M validation set.
Following our controllable setting, for each frame we first identify HOI classes touched by the control trajectories and compute class-wise IoU on those classes only.
We then report a dataset-level weighted metric, denoted as $m\mathrm{IoU}_w$, where each class contribution is weighted by its total ground-truth area:
$m\mathrm{IoU}_w=\sum_{c\in\mathcal{C}_{\mathrm{ctrl}}}\omega_c\,\mathrm{IoU}_c,\ \ \omega_c=\frac{A_c^{\mathrm{gt}}}{\sum_{k\in\mathcal{C}_{\mathrm{ctrl}}}A_k^{\mathrm{gt}}}$, where $\mathcal{C}_{\mathrm{ctrl}}$ denotes trajectory-controlled classes and $A_c^{\mathrm{gt}}$ is the ground-truth pixel count of class $c$.
Our model obtains an $m\mathrm{IoU}_w$ of $0.71$.
Despite temporal and spatial inconsistencies in both pseudo ground-truth masks and augmentor predictions, this result indicates that the predicted HOI masks remain well aligned with the target mask distribution.

\section{Additional Improvement}
\label{sec:traj_overlay}
\begin{figure}
	\centering
	\includegraphics[width=\columnwidth]{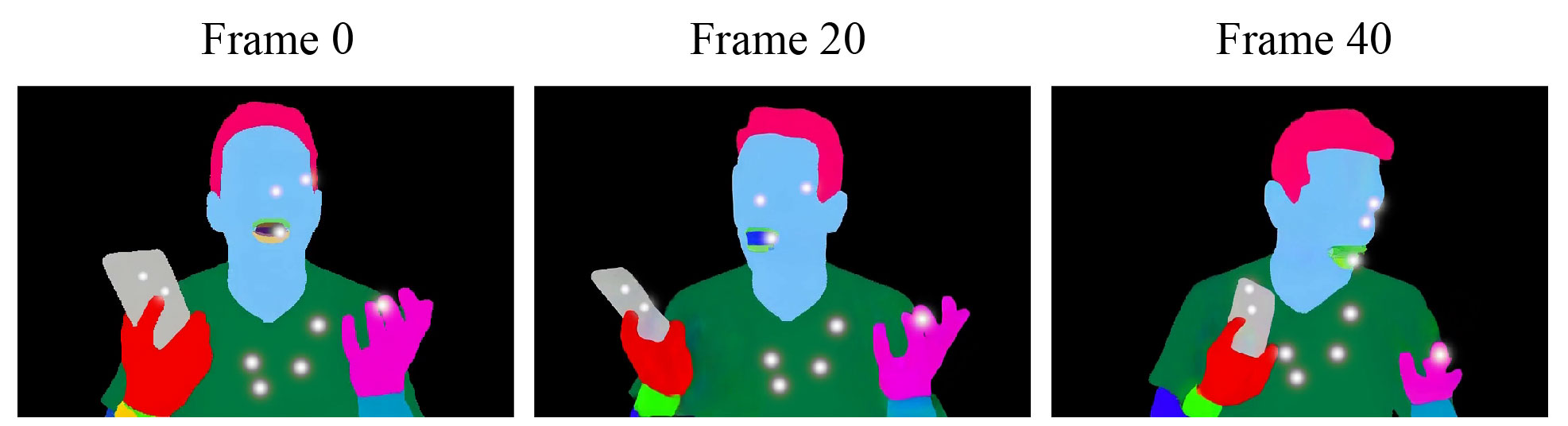}
	\caption{
		Overlaying the trajectory on the HOI masks improves trajectory adherence during generation, as the dense model leverages both the sparse trajectory input and the generated HOI masks. The figure shows three frames of inferred HOI masks during validation, with trajectories indicated by white dots.
        }
	\label{fig:traj_overlay}
\end{figure}

While \methodname{} significantly outperforms existing methods in terms of video quality, its comparable performance on TE and CA metrics may stem from the dense model’s lack of explicit trajectory awareness when structured HOI masks are used as the sole motion guidance. 
To address this, we introduce a simple modification: overlaying trajectories onto object masks during training and evaluate on overlaid HOI masks for validation, as illustrated in \cref{fig:traj_overlay}. This results in consistent improvements in both TE and CA, as reported in \cref{tab:traj_overlay}.
We also tried to additionally overlay the trajectories on the human segmentation masks during training, but we found that it did not lead to additional performance gain.
\begin{table}[hbt]
\centering
\resizebox{\linewidth}{!}
{
\begin{tabular}{llrrrrrrrrrr}
    \toprule
    & \multirow{2}{*}{Method} & \multirow{2}{*}{FVD$\downarrow$} & \multirow{2}{*}{TE$\downarrow$} & \multirow{2}{*}{CA$\uparrow$} & \multirow{2}{*}{CLIPSIM$\uparrow$} & \multicolumn{6}{c}{VBench} \\
    \cmidrule(lr){7-12}
    &  &  &  &  &  & SC$\uparrow$ & BC$\uparrow$ & DD$\uparrow$ & MS$\uparrow$ & AQ$\uparrow$ & IQ$\uparrow$ \\
     & \ours{\textbf{\methodname{}}}                                               &\best{915}  & 10.64     & 0.827 & 0.3036 & \best{0.93} & \best{0.94} & 0.58 & \best{0.99} & \best{0.51} & \best{0.68} \\ 
          & \ours{\textbf{\methodname{}-Traj}}                                               &933  & \best{8.74}      & \best{0.833} & \best{0.3042} & \best{0.93} & \best{0.94} & \best{0.62} & 0.98 & \best{0.51} & 0.67 \\          
    \bottomrule
\end{tabular}
}
\caption{
Overlaying trajectories onto object masks during dense model training further improves TE and CA on HOIGen dataset.}
\label{tab:traj_overlay}	
\end{table}

\section{HOI mask color palette}
\label{sec:color_palette}
\begin{figure}[!t]
	\includegraphics[width=0.8\linewidth]{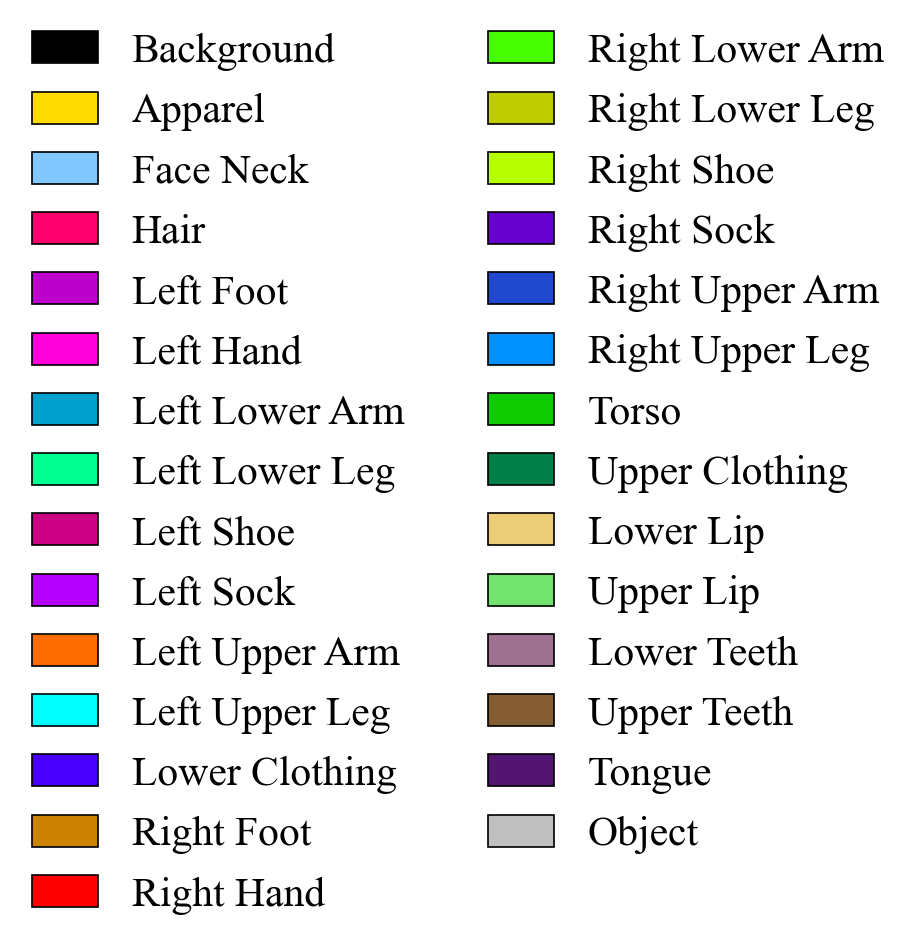}
	\caption{
        \textbf{HOI Masks Color Palette.}
		We visualize the color encoding of the 29 classes, where each color corresponds to a distinct part. The color scheme for human parts follows SAPIEN~\cite{khirodkar2024sapiens}, and light gray is used for the object.
        }
	\label{fig:color_palette}
\end{figure}

We visualize the HOI-mask color palette in \cref{fig:color_palette}. This palette introduces both part and interaction awareness into the augmentor, enabling richer motion guidance for the dense model.

\section{Augmentor prompt processing}
\label{sec:augmentor_prompt}
For training the augmentor, the text prompt needs to get rid of the appearance, background and lighting detail, to facilitate HOI mask generation.
We use Qwen3~\cite{qwen3} to distill a motion-centric prompt based on the original video caption.
Lastly, we append the fixed Sapien palette legend so that the generated prompt always refers to the canonical mask colors used by the augmentor.
\begin{figure}
    \centering
    \setlength{\fboxsep}{10pt}
    \setlength{\fboxrule}{0.4pt}
    \fcolorbox{black}{gray!5}{
    \begin{minipage}{0.95\columnwidth}
        \small\ttfamily
        Task: Rewrite the caption as a concise description of only the human and foreground object motion. \\
        Do not describe background, scene, appearance, clothing, colors, lighting, camera movement, or objects not interacted with. \\
        Focus only on the main action an use up to four short sentences. Output as one line without line breaks. \\
        Caption: ``\textit{Original video caption}'' \\
    \end{minipage}}
    \caption{Prompt template used to process text prompts for augmentor training. We prompt Qwen3~\cite{qwen3} to generate motion-centric captions aligned with the Sapien HOI-mask palette. We append the mask color scheme to the output: ``Use Sapien human mask colors (fixed palette): background black; object gray; face neck baby blue; hair hot pink; torso green; left hand magenta, right hand red; left lower arm teal, right lower arm neon green; left upper arm orange, right upper arm dark blue.''}
    \label{fig:augmentor_prompt}
\end{figure}

\end{document}